\documentclass[sigconf]{aamas} 

\usepackage{etoolbox}
\usepackage{xifthen}
\usepackage{xargs}
\usepackage[normalem]{ulem}
\usepackage{amsmath}
\usepackage{amsfonts}
\usepackage{mathtools}
\usepackage{bm}
\usepackage{color}
\usepackage{graphics} 
\usepackage{graphicx}
\usepackage{epsfig} 
\usepackage{epstopdf}
\usepackage{tikz}
\usetikzlibrary{calc,positioning,shapes,shadows,arrows,fit}
\usepackage{fancyhdr}
\usepackage{fancyref}
\usepackage{hyperref}
\usepackage{float}
\usepackage[font = footnotesize]{caption}
\usepackage{algorithm}
\usepackage{algorithmic}
\usepackage{caption}
\usepackage{subcaption}
\usepackage{url}
\makeatletter
\g@addto@macro{\UrlBreaks}{\UrlOrds}
\makeatother
\usepackage[inline, shortlabels]{enumitem}
\setlist{nolistsep, leftmargin=*}

\setcopyright{none} 
\renewcommand\footnotetextcopyrightpermission[1]{} 
\title{Symbolic Reinforcement Learning for Safe RAN Control}

\author{Alexandros Nikou}
\affiliation{
  \institution{Ericsson Research}
    }

\author{Anusha Mujumdar}
\affiliation{
  \institution{Ericsson Research}
    }
  
\author{Marin Orli\'{c}}
\affiliation{
  \institution{Ericsson Research}
    }
    
\author{Aneta Vulgarakis Feljan}
\affiliation{
  \institution{Ericsson Research}
    }

\begin{abstract} 
In this paper, we demonstrate a Symbolic Reinforcement Learning (SRL) architecture for safe control in Radio Access Network (RAN) applications. In our automated tool, a user can select a high-level safety specifications expressed in Linear Temporal Logic (LTL) to shield an RL agent running in a given cellular network with aim of optimizing network performance, as measured through certain Key Performance Indicators (KPIs). In the proposed architecture, network safety shielding is ensured through model-checking techniques over combined discrete system models (automata) that are abstracted through reinforcement learning. We demonstrate the user interface (UI) helping the user set intent specifications to the architecture and inspect the difference in allowed and blocked actions.
\end{abstract}

\newcommandx{\fix}[2][1=]{  
	\relax
	\ifmmode
		\ifthenelse{\isempty {#1}}
		{{\color{blue}{#2}}}
		{{\cancel{#1}}~{\color{blue}{#2}}}
	\else
		\ifthenelse{\isempty {#1}}
		{{\color{blue}{#2}}}
		{{\sout{#1}} {\color{blue}{#2}}}
	\fi
						}

\begin{document}


\pagestyle{fancy}
\fancyhead{}


\maketitle 


\section{Introduction and Motivation}
\label{sec:Introduction}

Due to the growing complexity of modern cellular networks, network optimization and control constitutes one of the main challenges. It is desirable by the Mobile Network Operators (MNOs) that the configuration is adjusted automatically in order to ensure acceptable Quality of Service (QoS) to each user connected to the network. In such application, the goal is to optimize a set of network KPIs such as \emph{coverage}, \emph{quality} and \emph{capacity} and to guarantee that certain bounds of these KPIs are not violated (safety specifications). This optimization is performed by adjusting the vertical electrical tilt of each of the antennas of the given network, known in the literature as remote electrical tilt (RET) optimization problem \cite{guo2013spectral, razavi2010fuzzy, fan2014self, buenestado2016self}.

Reinforcement learning (RL) \cite{sutton2018reinforcement, mnih2015human, garcia2015comprehensive, bouton2020point} has become a powerful solution for dealing with the problem of optimal decision making for agents interacting with uncertain environments. 
However, it is known that the large-scale exploration performed by RL algorithms can sometimes take the system to unsafe states \cite{garcia2015comprehensive}. In the problem of RET optimization, RL has been proven to be an effective framework for KPI optimization due to its self-learning capabilities and adaptivity to potential environment changes \cite{vannella2020off}. For addressing the safety problem (i.e., to guarantee that the desired KPIs remain in specified bounds) authors in \cite{vannella2020off} have proposed a statistical approach to empirically evaluate the RET optimization in different baseline policies and in different worst-case scenarios. 

The aforementeioned statistical approach focus on ensuring the reward value remains above a desired baseline. However, more widely accepted notions of safety are expressed in terms of safe states, defined according to a (formal) intent specification \cite{fulton2018safe}. The approach in \cite{fulton2018safe} decouples the notion of safety from that of reward. Intuitively, safety intents define the boundaries within which the RL agent may be free to explore. Motivated by the abovementioned, in this work, we demonstrate a novel approach for guaranteeing safety in RET optimization problem by using model-checking techniques and in parallel, we seek to generalize the problem in order to facilitate richer specifications than safety. In order to express desired specifications to the network into consideration, LTL is used (see \cite{katoen, loizou_2004, alex_automatica_2017, alex_PhD}), due to the fact that it provides a powerful mathematical formalism for such purpose. Our proposed demonstration exhibits the following attributes:
\begin{enumerate}
\item a general automatic framework from LTL specification user input to the derivation of the policy that fulfills it; at the same time, blocking the control actions that violate the specification;
\item novel system dynamics abstraction to companions Markov Decision Processes (MDP) which is computationally efficient;
\item UI development that allows the user to graphically access all the steps of the proposed approach.
\end{enumerate}

\textbf{Related work}.  Authors in \cite{alshiekh2018safe} propose a safe RL approach through shielding. However, they assume that the system dynamics abstraction into an MDP is given, which is challenging in network applications that this demonstration refers to. As mentioned previously, authors in \cite{vannella2020off} address the safe RET optimization problem, but this approach relies on statistical guarantees and it cannot handle general LTL specifications that we treat with this manuscript.

\begin{figure*}
\centering
\includegraphics[width=12.0cm,height=7.0cm]{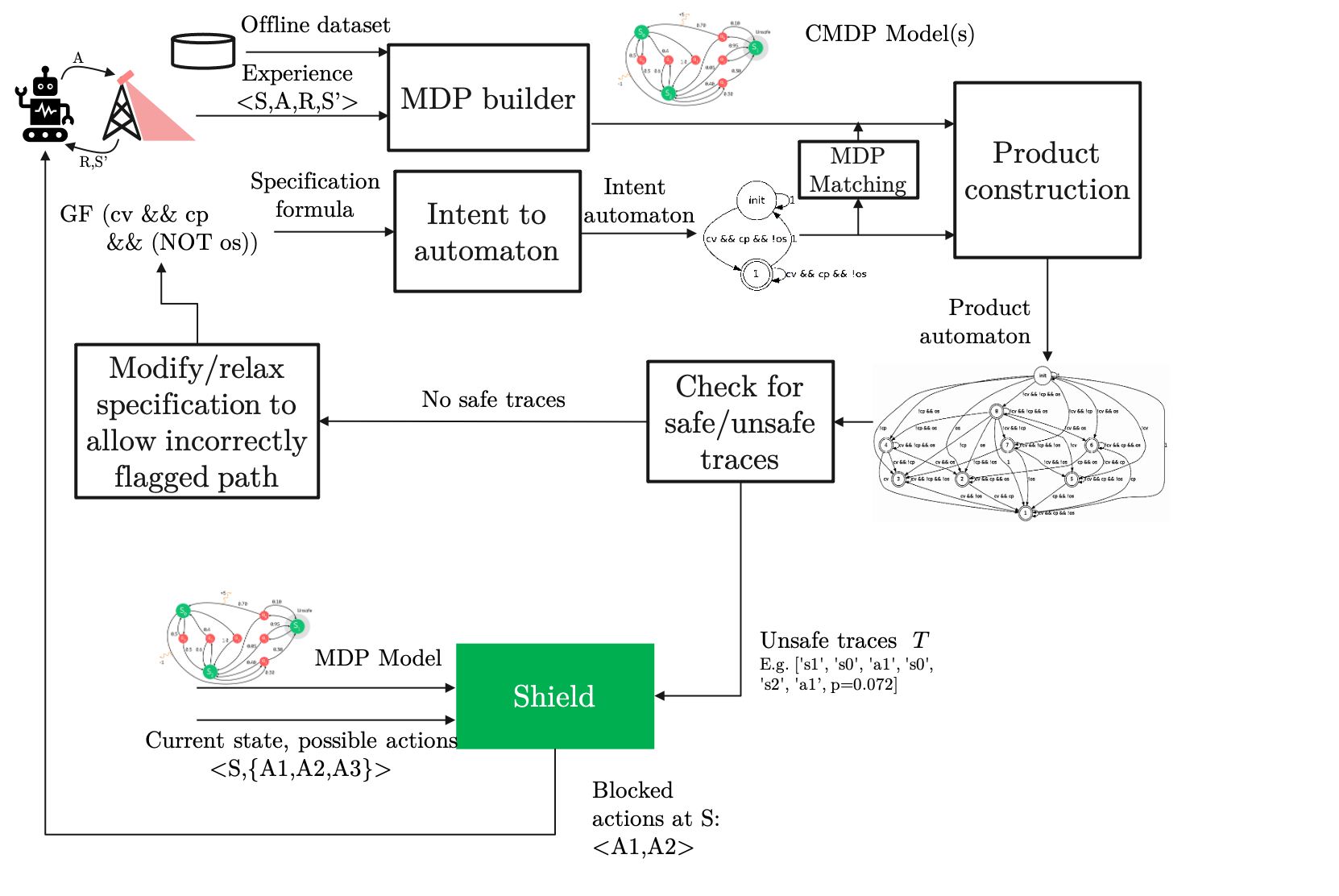}
\caption{A graphical illustration of the proposed architecture.}
\label{fig:test1}
\end{figure*}

\begin{figure*}
\centering
\includegraphics[width=15.0cm,height=6.0cm]{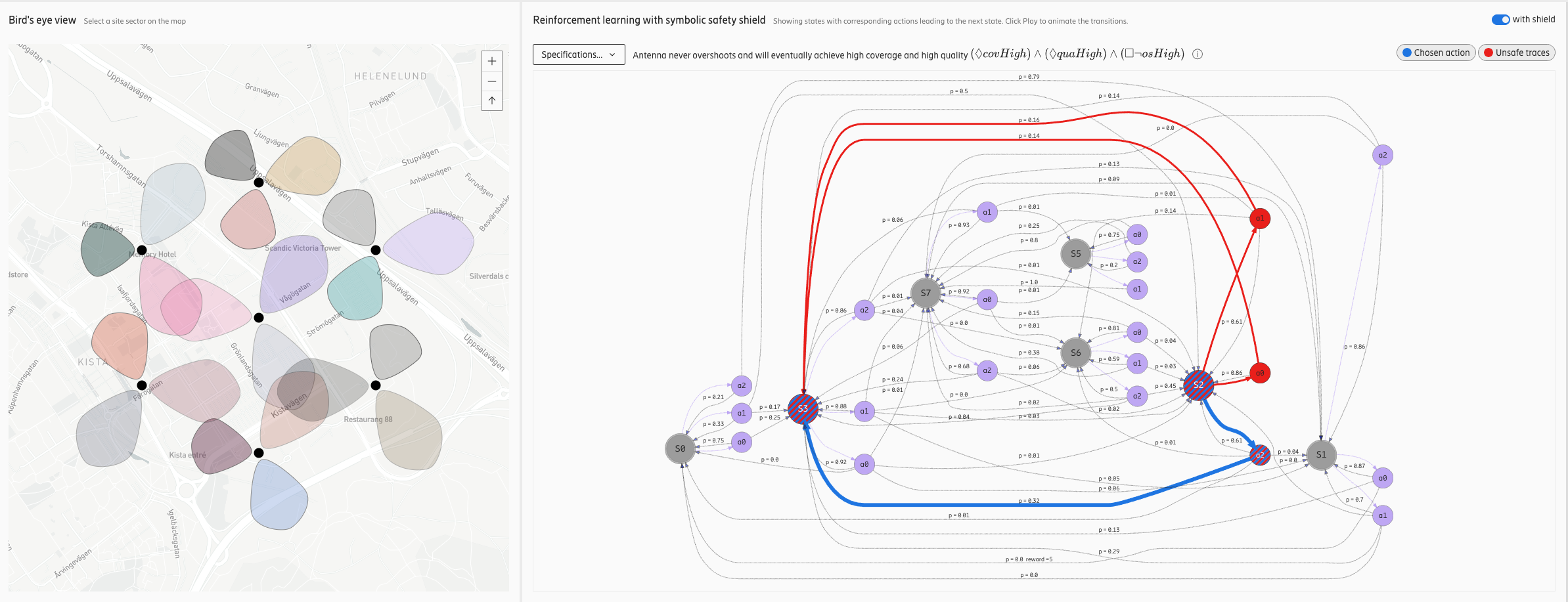}
\caption{A graphical illustration of the UI for the demonstration of the approach. The user can chose: the desired LTL formula, the resulting BA; and the evolution of the RL agent training and the actions blocked by the safety shield.}
\label{fig:UI}
\end{figure*}

\section{Demonstration}

Our key contribution is a proposed architecture which allows for intent specifications in RL, demonstrated in a real-world example. Here we focus on task specifications given in LTL. The syntax of LTL (see \cite{katoen}) over a set of atomic propositions $\Sigma$ is defined by the grammar $$\varphi := \top \ | \ \varsigma \ | \ \neg \varphi \ | \ \varphi_1 \wedge \varphi_2 \ | \ \bigcirc \varphi  \ | \  \varphi_1 \ \mathfrak{U} \ \varphi_2 \ | \ \Diamond \varphi \ | \ \square \varphi,$$ where $\varsigma \in \Sigma$ and $\bigcirc$, $\mathfrak{U}$, $\Diamond$, $\square$ operators stand for the next, until, eventually and always operators, respectively; $\neg$ and $\wedge$ are the negation and conjunction operator respectively. Every LTL formula can be translated to a B\"uchi Automaton (BA) that models all the system traces satisfying the formula \cite{gastin2001fast}.

Consider a geographic area covered by Radio Base Stations (RBS) and cells that serve a set of UEs uniformly distributed in that area. The RET optimization problem has goal to maximize network capacity and coverage while minimizing interference between the antennas. The RET control strategy handles the antenna tilt of each of the cells (agents), and is executed independently for each cell. The environment of the RL agents is a simulated mobile network as it can be seen in Fig. \ref{fig:UI}. The system dynamical model is captured through an MDP $(\mathcal{S}, \mathcal{A}, \mathcal{P}, \mathcal{R}, \gamma)$ where: $\mathcal{S}$ are the states consisting of values for down-tilt and KPIs (coverage, capacity and quality); actions $\mathcal{A} = \{\text{downtilt}, 0, \text{uptilt}\}$; transition probability matrix $\mathcal{P}$ which describes the state evolution given the current and the executed by the action state; rewards $\mathcal{R}$; and, discount factor $\gamma$. The RL agent's policy $\pi: \mathcal{S} \to \mathcal{A}$ is a function that maps the states to actions that define the agent's strategy. 

Our solution takes a sequence of steps to match the LTL specification with the RL agent as it is depicted in Fig. \ref{fig:test1} and block the actions that lead to unsafe states. Initially, the desired user specification is translated into LTL logic and subsequently into a BA. Then, by gathering the experience data tuples from the RL agent which is trained to a simulation environment with state-of-the-art model-free RL algorithms (DQN, Q-learning, SARSA \cite{guo2013spectral, razavi2010fuzzy, fan2014self, buenestado2016self}) we construct the system dynamics modelled as an MDP. In this solution, we have a novel structure known as Companion MDPs (CMDPs). CMDPs encode the state transitions only in terms of the subset of features specified in the intent, not the full set of state features.
This reduces the state space complexity, and keeps only the relevant features depending on the intent. An MDP matching component matches the intent to the relevant CMDP (depending on the features mentioned in the intent).

The experience data tuples that are generated during training are in the form $(s, a, r, s')$ where $s$ indicates the current state, $a$ the executed action, $r$ the received reward that the agent receives after applying action $a$ at state $s$; and $s'$ the state the agent transitions to after executing action $a$ at state $s$. In order to match the BA from the given LTL specification and the MDP, the states of the MDP are labelled according to the atomic propositions from the LTL specification. Then, by computing the product of the MDP with the specification, we construct: an automaton $\mathcal{A}_{\varphi}$ that models all the system behaviours over the given specification; an automaton $\mathcal{A}_{\neg \varphi}$ that models all the traces violating the specification. Then, by utilizing graph techniques and model checkers, we are able to find all the system traces violating the specification (w.r.t the trained MDP); if no safe traces are found from all states in the MDP, the user specification cannot be satisfied, which means that the LTL has to be modified (or relaxed). If there exist some unsafe and some safe traces, then the process moves to a shield strategy that blocks the actions that lead to unsafe traces. 
This process is depicted more formally in Fig. 1 and Algorithm 1. 

\section{Discussions}

\textbf{Interaction with the UI}. The UI is designed to be used by a network operations engineer who can specify safety intents, monitor tilts and their impact, and supervise the RL agent's operation. The initial screen of the UI depicts a geographic area with the available radio sites and cells. By clicking to one of the cells, a new screen appears with the KPI values depicted on the left. On the right part of the page, one can see: 1) the MDP system model; 2) a list of available LTL intents; 3) BAs representing each of the intents; 4) the button "Run safe RL" to run the simulation; and 5) the switch "with/without shield" for enabling the safety shield. The chosen actions on the MDP are depicted in blue, while the blocked actions by the shield are depicted in red. The user can view the training process and the optimal choice of actions that guarantee the satisfaction of given input as well as the block of unsafe actions. The current high level of detail in the UI is meant to illustrate the technology, it can be imagined that a production UI would instead show a summary of selected and blocked actions instead of large MDP models. The impact of the shield may also be viewed, and it is seen that the shield blocks a proportion of unsafe states (leading to 639 unsafe states instead of 994 without the shield). Interestingly, the shield also leads to a 68$\%$ improvement in the reward values. The video accompanying this paper can be found in:\\

\hspace{-5mm} \url{https://www.ericsson.com/en/reports-and-papers/research-papers/safe-ran-control} \\

\textbf{Applicability to other domains.} The proposed architecture is general and it can be applied to any framework in which the under consideration dynamical system is abstracted into an MDP, while LTL specifications need to be fulfilled. For example, in a robot planning applications, the states are locations of the environment that the robot can move, and atomic propositions are the goal state and the obstacles. The LTL formula of such application would include reachability and safety tasks. 

\textbf{Conclusions and future work.} In this paper, we have demonstrated an architecture for network KPIs optimization guided by user-defined intent specifications given in LTL. Our solution consists of MDP system dynamics abstraction, automata construction and products and model-checking techniques to block undesired actions that violate the specification. Future research directions will be devoted towards applying the proposed framework in other telecom use cases as well as in robotics (motion planning).

\textbf{Acknowledgements.}  The authors thank Ezeddin Al Hakim, Jaeseong Jeong, Maxime Bouton, Swarup Mohalik, Pooja Kashyap and Athanasios Karapantelakis for the fruitful discussion in topics related to this work and support in the simulation environment. Moreover, special thanks to Ericsson Research for supporting and funding our work.

\begin{algorithm}[t!]
\caption{}
\begin{algorithmic}[1]
\footnotesize{
\item[]\textbf{Input:} User specification $\varphi$
\STATE\textbf{Gather} experience replay $(s, a, r, s')$ from data;
\STATE\textbf{Discretize} states into $N_{b}$. State space size is $|\mathcal{S}|^{N_b}$;
\STATE\textbf{Construct} the MDP dynamics $(\mathcal{S}, \mathcal{A}, \mathcal{P}, \mathcal{R}, \gamma)$;
\STATE\textbf{Translate} the LTL formula $\Phi$ to a BA $\mathcal{A}_{\varphi}$;
\STATE\textbf{Compute} the product $\mathcal{T} = MDP \otimes \mathcal{A}_{\varphi}$ and pass it to model checker;
\STATE\textbf{Model checking} returns traces that violate $\varphi$;
\STATE\textbf{If} no safe traces found \textbf{Modify/Relax} $\varphi$
\STATE\textbf{Else} Block unsafe actions by function \textit{Shield}(MDP, $\mathcal{T}$).
}
\end{algorithmic} 
\end{algorithm}

\bibliographystyle{ACM-Reference-Format} 
\bibliography{references}
\end{document}